\documentclass{article}
\usepackage{blindtext}
\usepackage[a4paper, total={6in, 8in}]{geometry}

\usepackage[numbers]{natbib}
\usepackage{lmodern}

\usepackage{amsmath,amssymb}
\usepackage{rotating}
\usepackage{graphicx}
\usepackage{booktabs}
\usepackage{array}
\usepackage{url}
\usepackage{orcidlink}
\usepackage{xcolor}
\usepackage{cleveref}
\usepackage{caption}
\usepackage{subcaption} % pour subfigure + (a) (b)
\usepackage[dvipsnames]{xcolor}

\begin{document}
%\articletype{Paper} %	 e.g. Paper, Letter, Topical Review...

\normalsize
\title{Energy-Efficient Implementation of Spiking Recurrent Cells on FPGA}

\author{Pascal Harmeling, Florent De Geeter and Guillaume Drion}

%\affil{$^1$Montefiore Institute, University of Liege, Liege, Belgium}

%\affil{$^*$Author to whom any correspondence should be addressed.}
\maketitle

%\email{pascal.harmeling@uliege.be}

%\keywords{Spiking recurrent cell, FPGA, Neuromorphic Computing, Energy Efficiency, Hardware Acceleration.}

\begin{abstract}
Spiking Neural Networks (SNNs) can significantly reduce energy consumption compared to conventional artificial neural networks when spiking activity is sparse and the neuron model is hardware-friendly. However, biologically faithful models are often too costly for hardware implementations like FPGAs, whereas highly simplified models, such as the Leaky Integrate-and-Fire (LIF) neuron, sacrifice essential neuronal dynamics. In this work, we present an FPGA accelerator for an SNN utilizing the Spiking Recurrent Cell (SRC) model, which offers an intermediate level of biological plausibility and hardware efficiency between simple LIF and complex conductance-based models. The SRC model features continuous spikes and critical dynamical properties like the refractory period, yet remains mathematically simple enough for efficient FPGA deployment. To optimize SRC computation, we propose a set of mathematical simplifications such as piecewise-defined approximations that eliminate costly non-linear functions ($\tanh$, $\exp$) and using scaling to avoid floating-point arithmetic. We then integrate these units into a cohesive hardware architecture using a dedicated binding layer to permit modularity and scalability of the network VHDL architecture. To demonstrate this modularity, two complete networks, with respectively 1 and 4 SRC layers, were implemented and validated using Spiking Traces (SpT) derived from, respectively, the MNIST and Fashion-MNIST datasets. Weight matrices computed offline are stored directly in LUT-registers without retraining or hardware-specific adaptation to strictly evaluate the robustness of SRC. The reference implementation achieves $96.31\%$ accuracy on MNIST with a $220$-image SpT and a processing time of $1.7424\text{ ms}$ per digit. We further investigate accuracy-energy trade-offs by reducing the SpT length and quantizing synaptic weights down to $4\text{ bits}$, achieving $93.32\%$ accuracy at $0.492\text{ mJ}$ per digit ($55\text{ images}$, $5\text{-bit}$ weights) and $92.89\%$ at $0.394\text{ mJ}$ ($44\text{ images}$, $4\text{-bit}$ weights). These results demonstrate that SRC-based SNNs deliver competitive performance and low energy consumption while preserving richer neuronal dynamics than standard LIF models.
\end{abstract}

%----------------------------------------------------------------------------- Introduction 
%----------------------------------------------------------------------------- Introduction 
\section{Introduction}\label{sec:Introduction}
Artificial neural networks (ANNs) have undergone constant evolution over the past few decades, surpassing human performance in many areas. However, their energy consumption for learning and inference is increasing \cite{Emma}, which hinders their large-scale deployment in energy-constrained environments. On the other hand, biological brains only consume a few tens of watts of power to function, due to the sparse spiking nature of their computation.

Spiking neural networks (SNNs) have emerged as a low-power alternative to traditional ANNs, leveraging event-driven processing to emulate the efficiency of biological neural computations. SNNs have shown promising performance-consumption trade-offs, especially when deployed on dedicated hardware \cite{Dampfhoffer,Yan}. However, SNN energy consumption strongly depends on spiking model complexity, as well as on the sparsity of the emitted spikes \cite{Aliyev}. There exists a plethora of spiking neuron model types, ranging from simple Integrate-and-Fire (IF) models \cite{abbott1999lapicque, feng2001integrate} up to biologically-inspired conductance-based model that follows the Hodgkin--Huxley formalism \cite{HodgkinHuxley, skinner2006conductance}. Although the field of neuronal modeling is in constant evolution, SNNs mostly implement leaky integrate-and-fire (LIF) models and their variants, due to their computational simplicity. 

SNNs performance and power efficiency also strongly depend on the type of hardware being used. SNNs can be implemented on Central Processing Units (CPUs), similarly to classical ANNs, but these implementations bring little advantage as CPU architecture is poorly suited for sparse computations. In particular, the limited parallelism of CPU processing makes such implementations unable to deliver real-time performance for large-scale networks \cite{Wang}. Application-Specific Integrated Circuits (ASICs) can overcome this limitation \cite{Modaresi}. Full parallelization then becomes possible by tailoring the hardware architecture to the entire neural network. However, this advantage is offset by the inability to reprogram the device once it has been manufactured.

Field-Programmable Gate Arrays (FPGAs) provide a third alternative, allowing for programmable, energy-efficient implementation of a very large number of tasks in parallel. In addition, the integration of memory blocks (BRAM) and Digital Signal Processing (DSP) modules within FPGAs makes it possible to closely emulate the behavior of neural networks, while ensuring efficient learning and adaptation to the target tasks with minimal energy consumption \cite{Li}. 
This hardware approach, commonly referred to as an FPGA-based accelerator, constitutes a particularly relevant solution for the study of SNNs. FPGA implementations of SNNs using LIF models have emerged over the past few years (e.g., \citet{Carpegna,Carpegna2,Gupta,Han,He,Li,Li2}). LIF model simplicity allows for reduced hardware resources and computation time, but their reset mechanism and their purely linear dynamics deviate from biological behavior.

In this work, we introduce an FPGA implementation of an SNN based on the Spiking Recurrent Cell (SRC) model \cite{DeGeeter}. SRC neurons are derived from classical recurrent neural network (RNN) architectures and integrate spiking behavior directly into their state dynamics, eliminating the need for an explicit reset mechanism. Moreover, the SRC equations explicitly account for biological mechanisms such as the refractory period, which is rarely implemented in conventional LIF models, and incorporate a mixed feedback structure common to conductance-based models \cite{sepulchre2019control} and analog neurons \cite{mendolia2025neuromodulable}. Specifically, our main contributions are as follows:
\begin{itemize}
\item We propose a series of mathematical simplifications to the original SRC equations to significantly improve hardware implementation efficiency on FPGAs.
\item We evaluate the performance of both the original and simplified equations in software, demonstrating negligible accuracy loss due to the approximations.
\item We design a modular architecture featuring a dedicated binding layer to facilitate general-purpose, flexible deployment.
\item We demonstrate the modularity and robustness of the VHDL architecture by evaluating two distinct network configurations (a single-layer network of $100$ neurons versus a four-layer network of $100$ neurons per layer) on the MNIST and Fashion-MNIST datasets, achieved via direct transfer of software-trained weights without hardware-specific retraining.
\item We perform an exhaustive exploration of the accuracy-energy trade-offs by analyzing the impacts of synaptic weight quantization and reduced temporal input lengths.
\end{itemize}
We demonstrate that the proposed FPGA implementation of the SRC network achieves competitive performance compared to state-of-the-art LIF-based approaches in terms of both classification accuracy and energy efficiency.

\section{Related work}
\Cref{tab:Workcmp1,tab:Workcmp2} summarize the performance of FPGA implementations of LIF-based SNNs in terms of, e.g., computation time, energy consumption per processed MNIST image, and image recognition accuracy. \Cref{tab:Workcmp1} contains models with online learning rules embedded on the FPGA. \Cref{tab:Workcmp2} shows models for which learning is performed offline. In both cases, the image recognition rate remains high and is often above 90\%.

The solutions proposed by \citet{Gupta,Carpegna,Carpegna2,Li2} achieve image processing times below one millisecond. In all cases, the use of LIF-type neurons enables a simpler yet efficient implementation of the neural network.

 	\begin{table}[h]
			\caption{Comparison of FPGA-based SNN implementations (MNIST) - On-chip learning.}
			\label{tab:Workcmp1}
			\centering
			\begin{tabular}{@{}l c c c}
			\toprule
			Design          & Li et al. \cite{Li} & Gupta et al. \cite{Gupta} & He et al. \cite{He} \\
 			\midrule
			Year            &   2021              &   2020                    &      2021           \\
			Model           &  LIF                &     LIF                   &      LIF            \\
            Neurons         &  784-200-100-10     &   784-16                  &   784-2048-100      \\
            FPGA            &  VIRTEX-7           &    XC6VLX240T             &      Zynq7045       \\
            $freq.$ [MHz]   &  100                &    100                    &     100 (200)$^1$   \\
            Used Logic cells&  -                  &   79,468                  &       23180         \\
            Used DSP        &  -                  &    64                     &        8            \\
            Used BRAM       &  -                  &    16                     &       129           \\
			Weight (bits)   &  16                 &   24                      &                     \\
            $T$/img [ms]    &  3.15               &  0.5                      &     66.6 (33.3)$^1$ \\
            E/img   [mJ]    &  5.04               &   -                       &                     \\
            Accuracy [\%]   &  92.93              &   -                       &        93.00        \\
			\bottomrule
			\multicolumn{4}{l}{$^1$ Initial values tested at 200 MHz}\\
            \end{tabular}
            \vspace{2mm}
    \end{table}
            
 	\begin{table}[h]
			\caption{Comparison of FPGA-based SNN implementations (MNIST) - Offline learning.}
			\label{tab:Workcmp2}
			\centering
			\begin{tabular}{@{}l c c c c}
			\toprule
	          Design          & Carpegna et al. \cite{Carpegna} & Carpegna et al. \cite{Carpegna2} & Han et al.  \cite{Han} & Li et al. \cite{Li2} \\
 			\midrule
			Year            &  2022                           &  2024                            &  2020                  &     2023            \\
			Model           &  LIF                            &  LIF                             &    LIF                 &    IF/LIF           \\
            Neurons         &  1384                           &  784-128-10                      &   784-1024-1024-10     &   784-16C3-..-10$^1$\\
            FPGA            &  ARTIX-7                        &  XC7Z020                         &  XC7Z045 SoC           &   XCZU3eg           \\
            $freq.$ [MHz]   &  100                            &  100                             &     100$^2$            &  300                \\
            Used Logic cells&  -                              &  7,612                           &    12,690              &  -                  \\
            Used DSP        &  -                              &  0                               &    -                   &  -                  \\
            Used BRAM       &  45                             &  18                              &   40.5                 &  -                  \\
			Weight (bits)   &  5                              &  4                               &    16                  &  -                  \\
            $T$/img [ms]    &  0.215                          &  0.78                            &   12.42$^2$            &  0.491              \\
            E/img   [mJ]    &  13                             &  0.14                            &    5.92$^2$            &  1.25               \\
            Accuracy[\%]  &  73.96                            &  93.85                           &    97.06               &  98.12              \\
			\bottomrule
            \multicolumn{5}{l}{$^1$ 784-16c3-64c3-p2-128c3-p2-256c3-256c3-10}\\
            \multicolumn{5}{l}{$^2$ adjust from 150 to 100 MHz  } \\
			\end{tabular}
            \vspace{2mm}
            \raggedright
 	\end{table}

The work by \citet{Carpegna} presents a layered architecture and an associated implementation method, achieving low resource utilization and reduced energy consumption. The study of \citet{Han} highlights that excessive quantization of the weights $W$ leads to a significant degradation in accuracy, suggesting the existence of a minimum word-length threshold. However, depending on the learning method, good performance can still be achieved with a 4-bit weight representation \cite{Panchapakesan}. In this work, we explore a wide quantization range, from 9 down to 2 bits (including the sign bit).

Spike management and inter-layer data transfer must also be carefully designed. The \textit{event-driven} approach proposed by \citet{Roy} is highly efficient when spikes are sparse, but becomes less effective under dense or bursty activity. In such cases, synchronous layer-by-layer processing \cite{Wu} is commonly adopted. Here, we employ a synchronous per-layer processing scheme, leveraging the parallelism offered by the FPGA.

\section{Adaptation of SRC equations for efficient FPGA implementation}\label{sec:Mathsimplications}
In our implementation, we use the SRC model in the intermediate layer to preserve continuous neuronal dynamics while remaining compatible with efficient hardware simplification. We first describe original SRC equations and then propose a set of mathematical simplifications that allow for efficient FPGA implementation. 

\subsection{Original SRC equations}
The SRC equations are written as

\begin{align}
			h[t] &=  \tanh (I[t] + r \cdot h[t-1] + r_{s} \cdot h_{s}[t-1] + b_{h}), \label{eq:src_model1} \\
			h_{s}[t] &= z_{s}[t] \cdot h_{s}[t-1] + (1- z_{s}[t]) \cdot h[t-1],\label{eq:src_model2} \\
             z_{s}[t] &= z^{hyp}_{s} - (z^{hyp}_{s} - z^{deep}_{s}) \frac{1}{1 + exp (-10 (h[t-1]-0.5))}.
             \label{eq:src_model3}
\end{align}

where $I[t]$ represents the current input to the system and $h[t]$ corresponds to its output (the spike-generation equation). The equation for $h_{s}[t]$ introduces refractory-type dynamics whose relaxation time is defined by $z_{s}[t]$. Higher values of $z_{s}[t]<1$ lead to slower relaxation, hence lower firing frequencies. The parameters $r$, $r_{s}$, $b_{h}$, $z^{hyp}_{s}$, and $z^{deep}_{s}$ are fixed and depend on the implementation method and the desired sensitivity. The values used by \citet{DeGeeter} are respectively $r=2$, $r_{s}=-7$, $b_{h}=-6$, $z^{hyp}_{s}=0.9$, and $z^{deep}_{s}=0.0$.

This study primarily focuses on the benefits of a hardware implementation on an FPGA. The learning process is deliberately only briefly addressed, since the parameters (bias $b_{h}$ and weights $w_{i,j}$) are directly taken from the network implemented in Julia. No transformation is required, which represents a significant advantage when porting the design to the VHDL language. The VHDL language was chosen over high-level synthesis tools (e.g., Vitis HLS or C language) to ensure full control over the hardware resources and to maximize implementation-level optimization.

%----------------------------------------------------------------------------- Math simplications
%----------------------------------------------------------------------------- Math simplications

\subsection{Simplified SRC equations}

The original SRC equations presented in \cref{eq:src_model1,eq:src_model2,eq:src_model3} are not directly suitable for efficient FPGA implementation, since they involve floating-point representations, nonlinear functions such as $\tanh(x)$ and $\exp(x)$, as well as multiplication operations. These computational elements result in excessive consumption of logic resources, DSP blocks, and memory, and also introduce substantial computational latency. This section derives mathematical simplifications that preserve the temporal dynamics of SRCs, but limit the implementation to simple operations, such as additions, bit shifts, and integer-based computations. We first remove floating-point arithmetic by scaling all variables and constant terms ($b_h$) by a factor of 1000, and represent them as integer (\texttt{Int32}). Coefficients ($r,r_s$) where also represented as integer without scaling. This integer representation provides sufficient numerical precision while keeping the internal variables within a compact signed range, with an effective range compatible with the interval $[-1024, +1023]$. The remaining simplification process is detailed below.

\begin{itemize}
    \item \textbf{Simplification of $h[t]$ (\cref{eq:src_model1}):}
     The first simplification consists in replacing the coefficient $r_{s}=-7$ by the even value $r_{s}=-8$, allowing to factor out 2 from the right-hand side of \cref{eq:src_model1}, and multiplying the constant term $b_h$ by 1000 ($b_h = 2\cdot3000)$. The original equation becomes 
     \begin{align}
        h[t] = 1000 \cdot \tanh (I[t] + 2 \cdot (h[t-1] - 4 \cdot h_{s}[t-1] - 3000)).\nonumber
     \end{align}
    
     This factorization allows to replace the multiplication by 2 by a left shift ($<<1$) and the multiplication by 4 by two left shifts ($<<2$), which gives
      \begin{align}
        h[t] = 1000\cdot \tanh (I[t] +(h[t-1] - (h_{s}[t-1]<<2) - 3000)<<1).
     \end{align}
     
     Replacing multiplications by shifts allows to replace mathematical operations with logical ones within the register (LUT). This approximation slightly modifies the excitability and refractory period behavior of the neuron. These two effects can be compensated for by correcting the minimum and maximum values of $z_{s}[t]$. Accordingly, the two boundary values $z^{deep}$ and $z^{hyp}$ are adjusted.

     In addition, the unary operator $tanh(x)$ is replaced by a piecewise-defined function $PWF(x)$ to simplify the number of operations      (\cref{fig:tanh}):
      \begin{align}
        PWF(x) = 
         \begin{cases}
            -1000 , \quad \text{if } ((2 \cdot x + x) / 4)<-1000\\
            +1000, \quad \text{if } ((2 \cdot x + x) / 4)>1000\\
             (2 \cdot x + x) / 4, \text{ otherwise},
         \end{cases}\nonumber
     \end{align}
     which we rewrite as
     \begin{align*}
        PWF(x) = \text{clamp}((2 \cdot x + x) / 4,-1000,+1000)
     \end{align*}
     The slope of the linear region was optimized to minimize the approximation error. Replacing multiplication by 2 by a left shift ($<<1$), and division by four by two right shifts ($>>2$) provides the final simplified equation
     \begin{align}
     x[t] &= I[t] + \left( h[t-1] - \left( h_s[t-1] << 2 \right) - 3000 \right) << 1,\nonumber \\
    h[t] &= \text{clamp}(((x[t]<<1)+ x[t])>>2,-1000,+1000).
     \end{align}

    \item \textbf{Simplification of $h_s[t]$ (\cref{eq:src_model2}):}
    We expand the product term $ (1- z_{s}[t]) \cdot h[t-1]$ and subsequently factor out $z_{s}$ from the right-hand side of \cref{eq:src_model2}. This allows to reduce the number of multiplications. To maintain the scaling by 1000, the term $z_{s}[t] \cdot(h_{s}[t-1] - h[t-1])$ must be divided by 1000, which we approximate by ten shifts to the right for efficiency ($>>10$). This corresponds to a division by $2^{10}$=1024. The resulting approximation error, approximately 2.4\%, is compensated for by adjusting the value of $z_s^{hyp}$. The final simplified equation becomes
    \begin{align}
    h_s[t] &= \left( z_s[t] \cdot \left( h_s[t-1] - h[t-1] \right) \right) >> 10 + h[t-1].
    \end{align}

    \item \textbf{Simplification of $z_s[t]$ (\cref{eq:src_model3}):}
    The operator $\exp(x)$ used in \cref{eq:src_model3}) cannot be easily simplified, and its use within a fraction further increases the complexity. However, one can observe that $z_{s}[t]$ acts as a switching mechanism between a slow and a fast dynamical regime, as shown in \cref{fig:exp}. In practice, the values of $z_{s}[t]$ remain close to either $z^{deep}_{s}$ or $z^{hyp}_{s}$. By determining a switching threshold, the function determining $z_{s}[t]$ can be replaced by a simple threshold rule: if $h[t-1]$ is lower than $V_{th}$, then $z_{s}[t]=z^{hyp}_{s}$; otherwise, $z_{s}[t]=z^{deep}_{s}$. the $V_{th}$ value is set experimentally to 500. The final simplified equation becomes
    \begin{align}
    z_s[t] &=
        \begin{cases}
            z_s^{hyp}, \quad \text{if }h[t-1] < 500,\\
            z_s^{deep} \text{ otherwise.}
        \end{cases}
    \end{align}
    This simplification slightly modifies the shape of the spike through slight changes in the timescale separation during spiking. 
    
   \begin{figure}[h]
    \centering
    \begin{subfigure}[c]{0.50\linewidth} % ou mettre b pour bottom
        \centering
        \includegraphics[width=7.4cm]{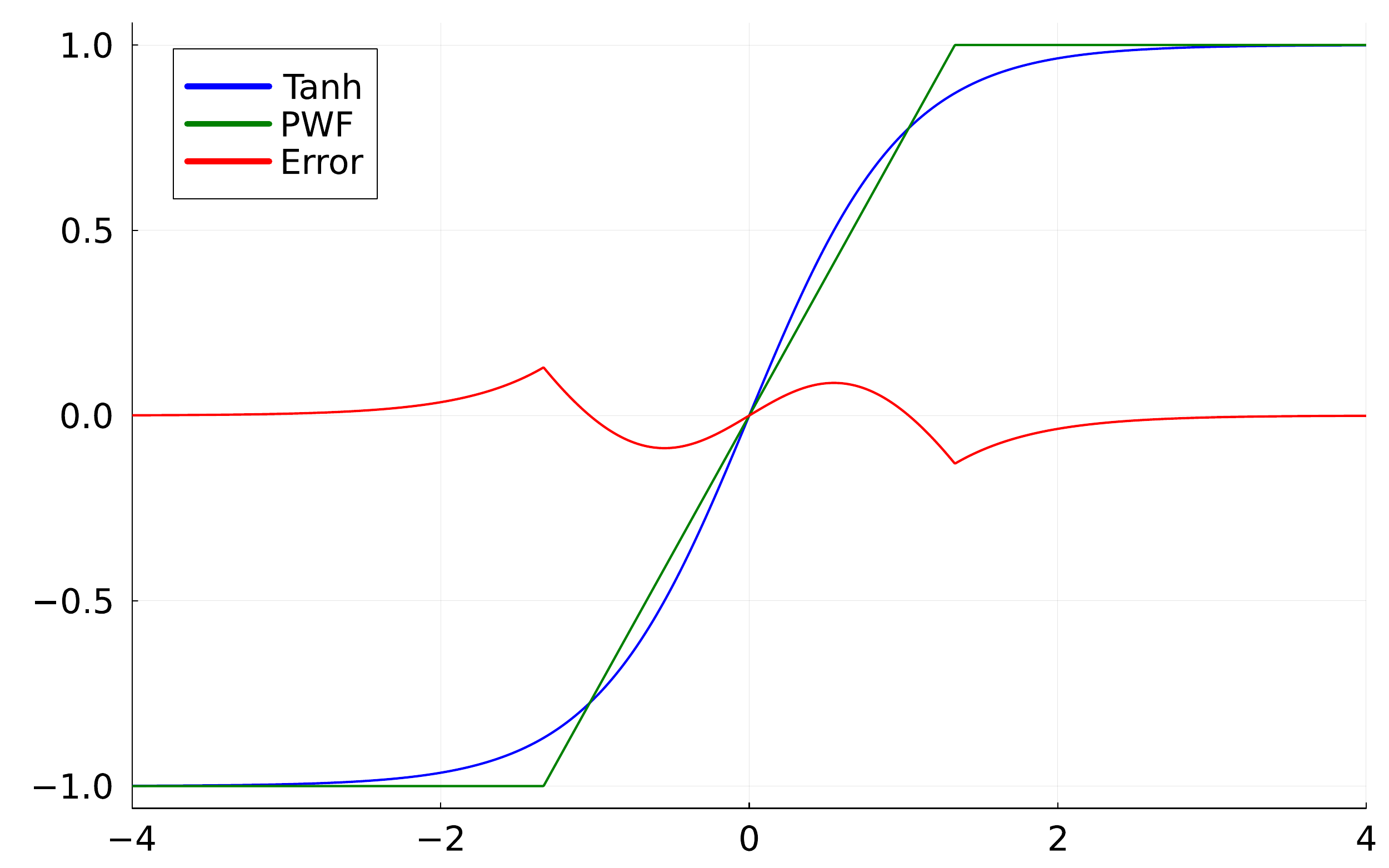}
        \caption{}
        \label{fig:tanh}
    \end{subfigure}\hfill
    \begin{subfigure}[c]{0.50\linewidth} % ou mettre b pour bottom
        \centering
        \includegraphics[width=5.5cm]{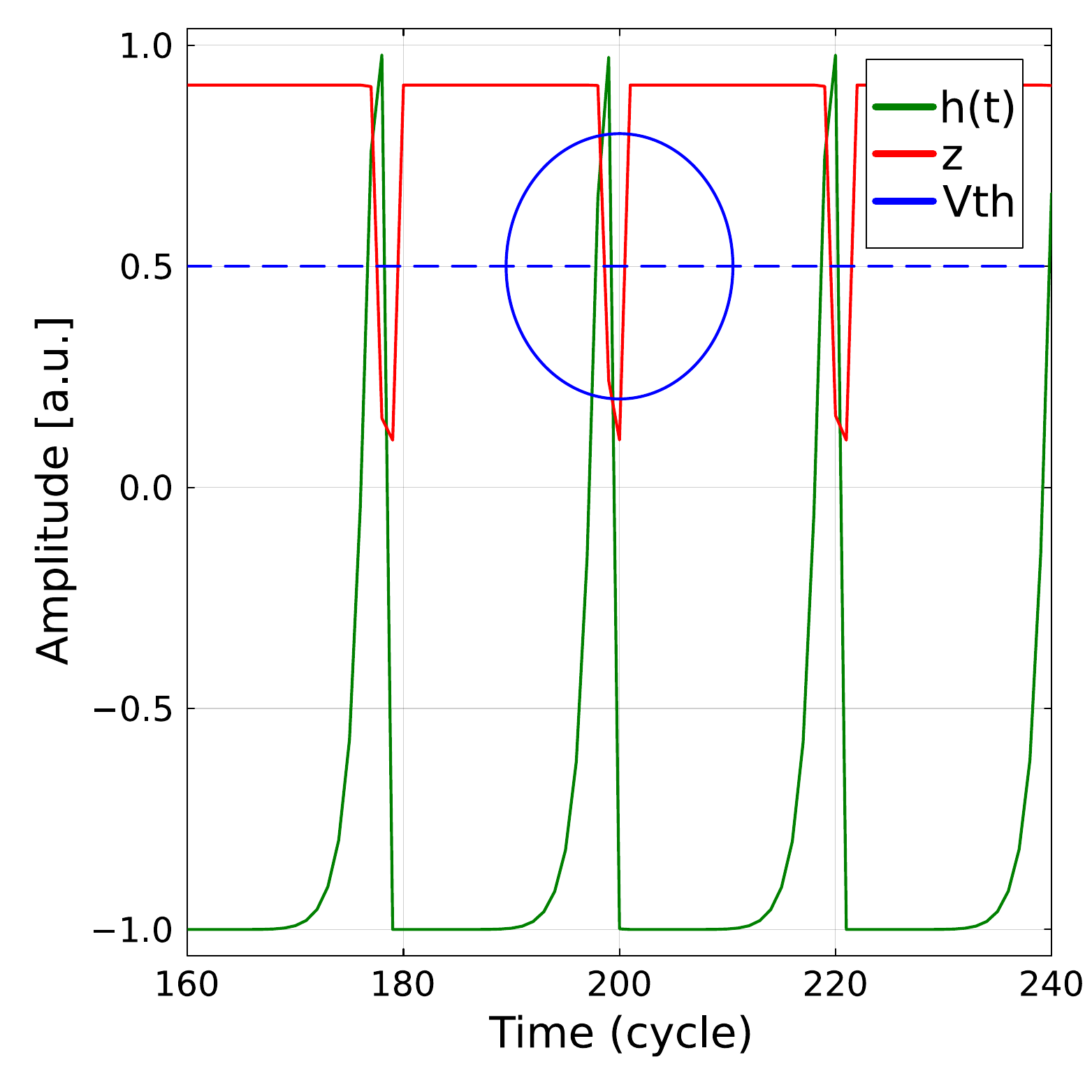}
        \caption{}
        \label{fig:exp}
    \end{subfigure}

    \caption{\textbf{(a)} Simplification of the $tanh$ function \textbf{(b)} Simplification of the function $z[t]$ into a threshold rule with detection threshold $V_{th}$.}
    \label{fig:HsvsZdeep}
\end{figure}
        
\end{itemize}
As a summary, the original SRC equations
\begin{align}
            x[t] &= I[t] + 2 \cdot h[t-1] - 7 \cdot h_{s}[t-1] + 6),\\
			h[t] &=  \tanh(x[t]), \nonumber \\
			h_{s}[t] &= z_{s}[t] \cdot h_{s}[t-1] + (1- z_{s}[t]) \cdot h[t-1],\nonumber \\
             z_{s}[t] &= z^{hyp}_{s} - (z^{hyp}_{s} - z^{deep}_{s}) \frac{1}{1 + \exp (-10 (h[t-1]-0.5))}.
             \nonumber
\end{align}
become
\begin{align}
    x[t] &= I[t] + \left( h[t-1] - \left( h_s[t-1] << 2 \right) - 3000 \right) << 1,\nonumber \\
    h[t] &= 
       \text{clamp}(((x[t]<<1)+ x[t])>>2,-1000,+1000), \nonumber
         \\
    h_s[t] &= \left( z_s[t] \cdot \left( h_s[t-1] - h[t-1] \right) \right) >> 10 + h[t-1] \nonumber \\
    z_s[t] &=
        \begin{cases}
            z_s^{hyp},\quad \text{if }  h[t-1] < 500,\\
            z_s^{deep}, \quad \text{otherwise.}
        \end{cases}\nonumber
\label{eqn:SRC_pseudo_vhdl_complete}
\end{align}

This formulation highlights the main advantage of the proposed simplification strategy. The original SRC equations are transformed into a sequence of operations that are naturally supported by FPGA logic: additions, subtractions, comparisons, shifts, and a limited number of multiplications. As a result, the simplified neuron preserves the main dynamical properties of the original SRC model while being much more suitable for an efficient VHDL implementation, and permits to avoid the use of DSPs.

\subsection{Comparison between original and simplified SRC equations}

\begin{figure}[h]
	\centering
	\includegraphics[width=14.5cm]{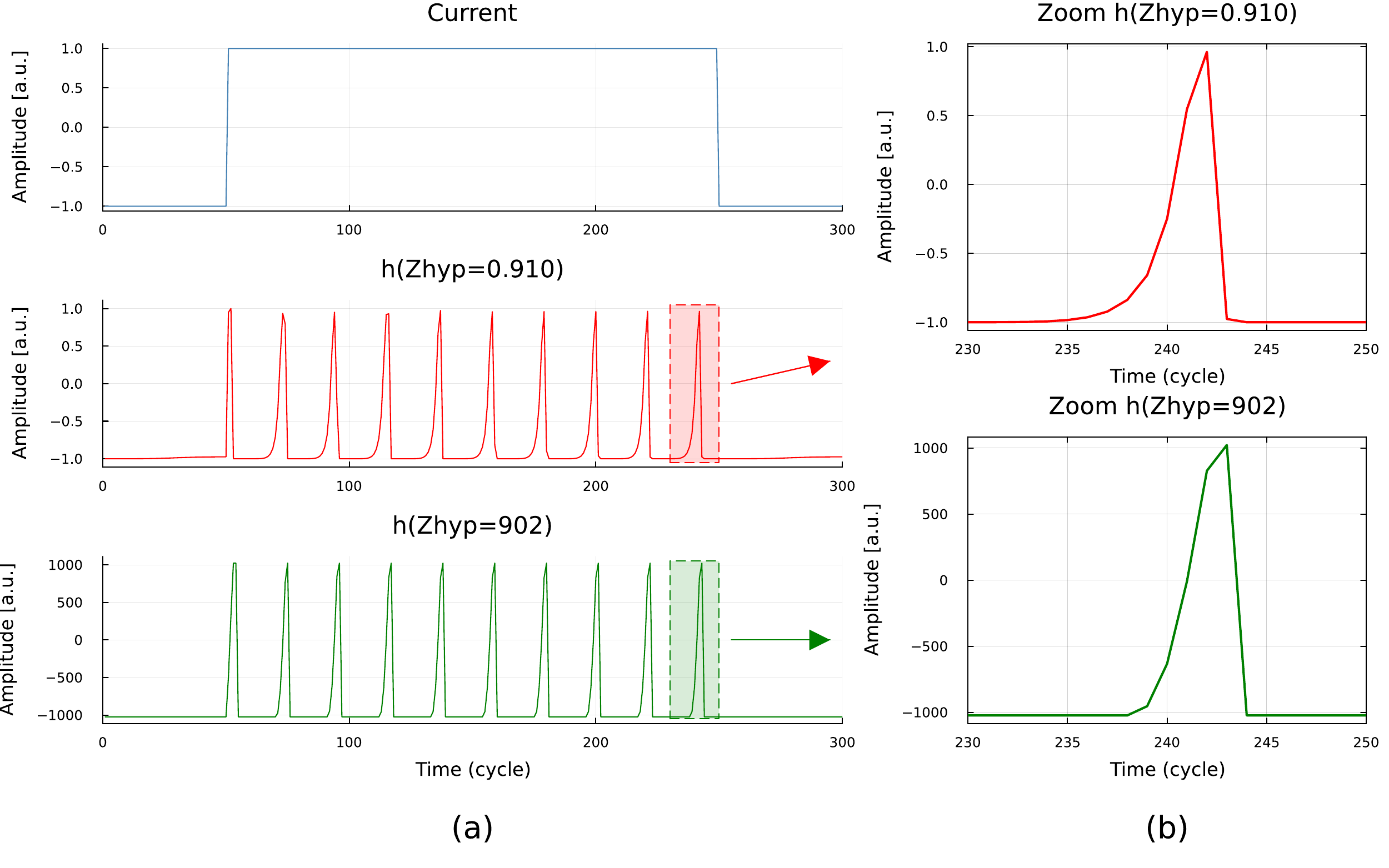}
	  \caption{\textbf{(a)} Response $h[t]$ of the original implementation (red) and PsV (green) to a same input current. \textbf{(b)} Spike shapes produced by the original (red) and PsV (green) implementations.}
    \label{fig:Table-julVHD}
\end{figure}

We next investigate the impact of these simplifications on model dynamics by comparing the outputs of the two different SRC implementations. The first version implements the original SRC equations in Julia using a \texttt{Float32} representation. The second version uses an \texttt{Int32} representation together with all the simplifications proposed above to satisfy the constraints of a VHDL implementation. This version is also implemented in Julia and called  Pseudo-VHDL (PsV). During this evaluation step, $z^{hyp}_{s}$ and $z^{deep}_{s}$ are initialized to 0.910 and 0.000 in the original implementation, and to 902 and 100 in the PsV implementation. The values 902 and 100 are chosen so as to minimize the discrepancy between the firing-frequency response of the \texttt{Float32} reference implementation and that of the pseudo-VHDL implementation. \Cref{fig:Table-julVHD} illustrates the behavior of the output $h[t]$ when implemented in Julia (red) and in PsV (green) when submitted to the same input current, showing a close agreement between the two versions.

\begin{figure*}[!t]
		\centering
		\includegraphics[width=14.5cm]{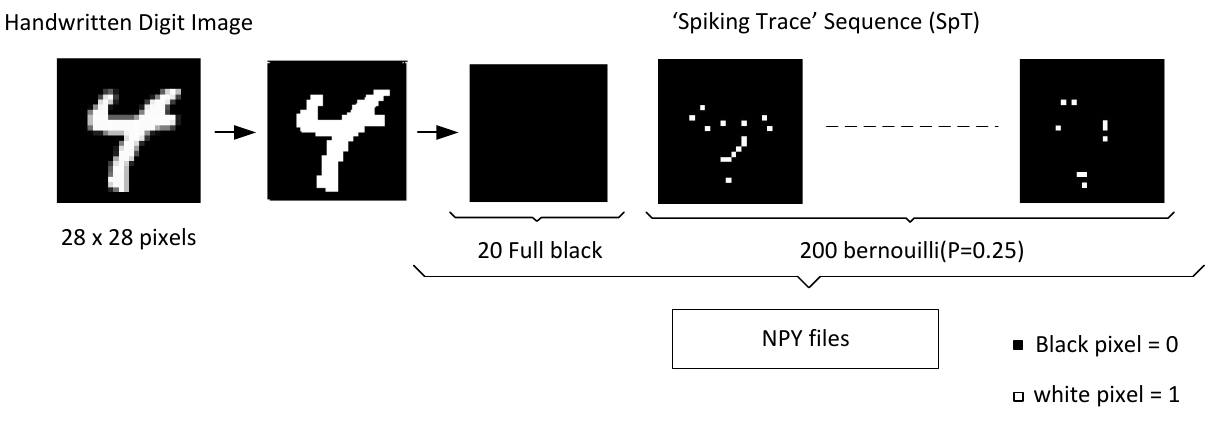}
		\caption{MNIST database processing to generate the SpTs applied to the SRC inputs layer. Each MNIST image is discretized into binary values (0/1). This binary image is used to generate a sequence of 220 images using a Bernoulli process with a maximum probability of 0.25, and 20 all-black images are added at the beginning, serving as a reset window. It results in a sequence of 220 images constituting the SpT (a white pixel corresponds to a spike). Each of the 220 images is sequentially fed to the SRC input layer as a vector of size 784.}
		\label{fig:SpikingTrace}
	\end{figure*}

We then compared inferences of the complete PsV network implementation on MNIST in Julia in order to compare the original SRC network with the simplified version. To ensure a fair comparison, it is essential to use the exact same Spiking Trace (SpT) sequences. A SpT is a sequence of binary images (0/1) intended to simulate a stream of spikes at the input. A SpT test set is constructed from a single MNIST sample, whose elements are processed following the procedure illustrated in \cref{fig:SpikingTrace}. The first step consists in discretizing an MNIST image into binary values (0/1): pixels with values greater than 0.5 are set to 1, otherwise to 0. This binary image is then used to generate a sequence of 200 images using a Bernoulli process with a maximum probability of 0.25. A filtering step prevents a given pixel from remaining at 1 in two consecutive images. Finally, 20 all-black images are added at the beginning, serving as a reset window as done by  \citet{DeGeeter}. The resulting sequence of 220 images constitutes a SpT. This processing is applied to all 10,000 images of the MNIST test set and saved into 10,000 NPY files. Over the entire test set, the original implementation achieves an accuracy of 96.48\%, while the PsV implementation reaches 96.31\%.

\begin{figure}[h!]
    \centering
    \begin{subfigure}[c]{0.50\linewidth} % ou mettre b pour bottom
        \centering
        \includegraphics[width=\linewidth]{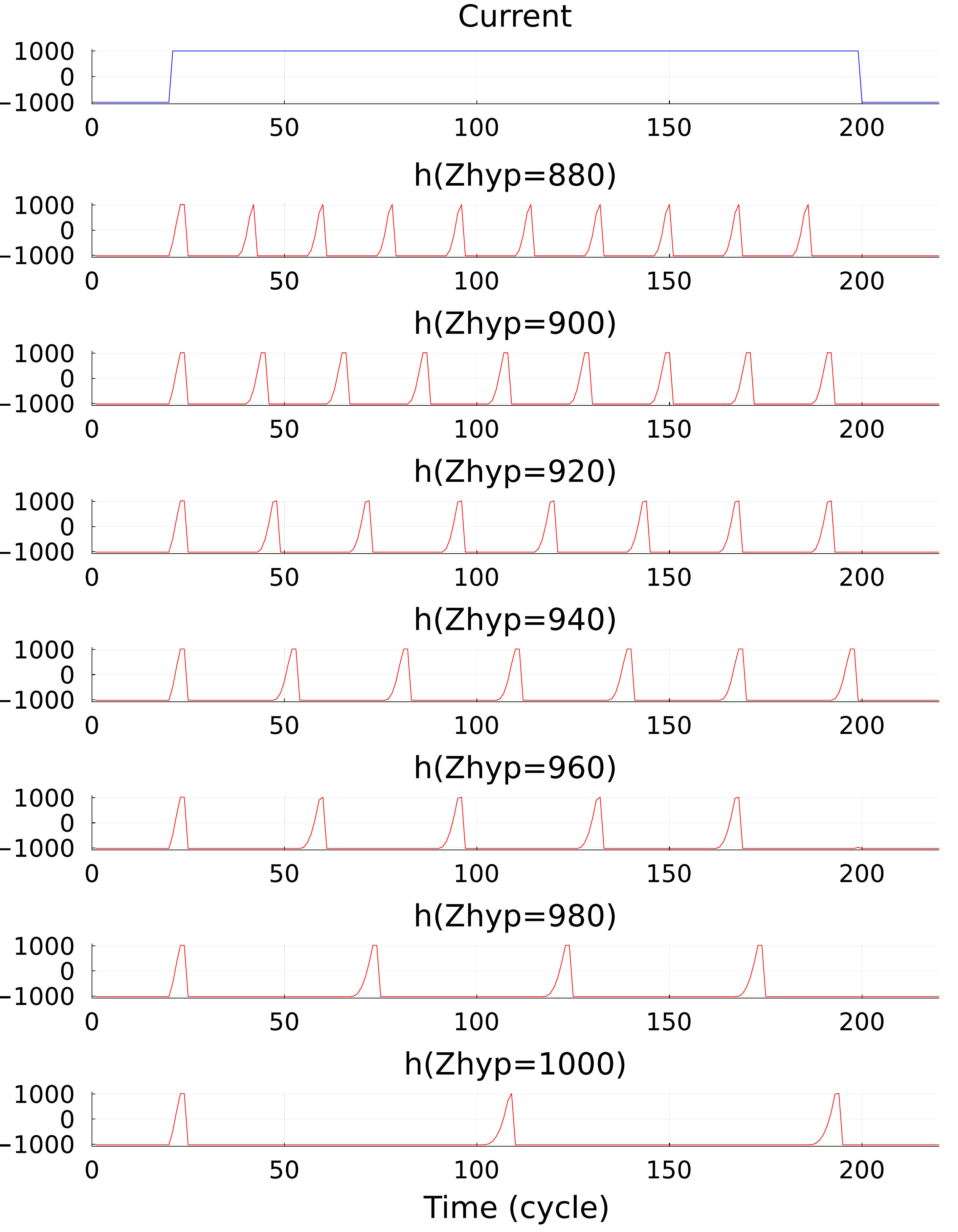}
        \caption{}
        \label{fig:HsvsZdeep_a}
    \end{subfigure}\hfill
    \begin{subfigure}[c]{0.5\linewidth} % ou mettre b pour bottom
        \centering
        \includegraphics[width=\linewidth]{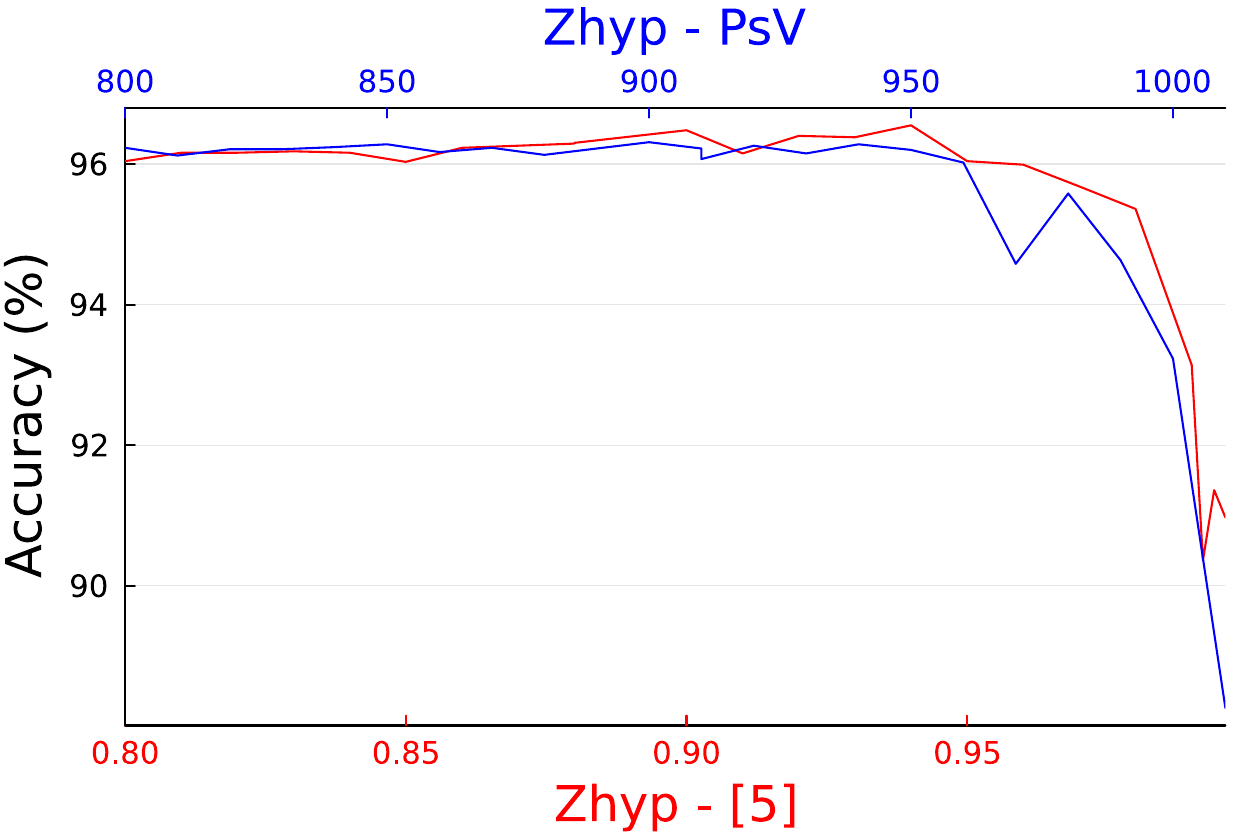}
        \caption{}
        \label{fig:HsvsZdeep_b}
    \end{subfigure}

    \caption{\textbf{(a)} Output spike frequency of the SRC neuron, governed by the $h_{s}$ equation, as a function of $z^{hyp}_{s}$. \textbf{(b)} Accuracy of the neural networks output results.}
    \label{fig:HsvsZdeep}
\end{figure}

To test performance stability with respect to spike frequency, a second series of tests is performed by modifying the value of the parameter $z^{hyp}_{s}$. \Cref{fig:HsvsZdeep_a} shows spiking frequency as a function of $z^{hyp}_{s}$. Approximately three times fewer spikes are observed when $z^{hyp}_{s}$ increases from 880 to 980.
\Cref{fig:HsvsZdeep_b} shows the accuracy of both models as a function of $z^{hyp}_{s}$. Both models show a high and stable accuracy for a large range of parameter values. This shows that it is possible to reduce the spiking rate at the output of the SRC neurons without significantly affecting accuracy, thereby enabling a reduction in energy consumption through increased spike sparsity. The range $z^{hyp}_{s}\in[880,980]$ will later be used to evaluate the stability and accuracy of the SRC network when implemented in VHDL and executed on an FPGA.

\begin{figure}[!t]
    \centering
    \includegraphics[width=15.5cm]{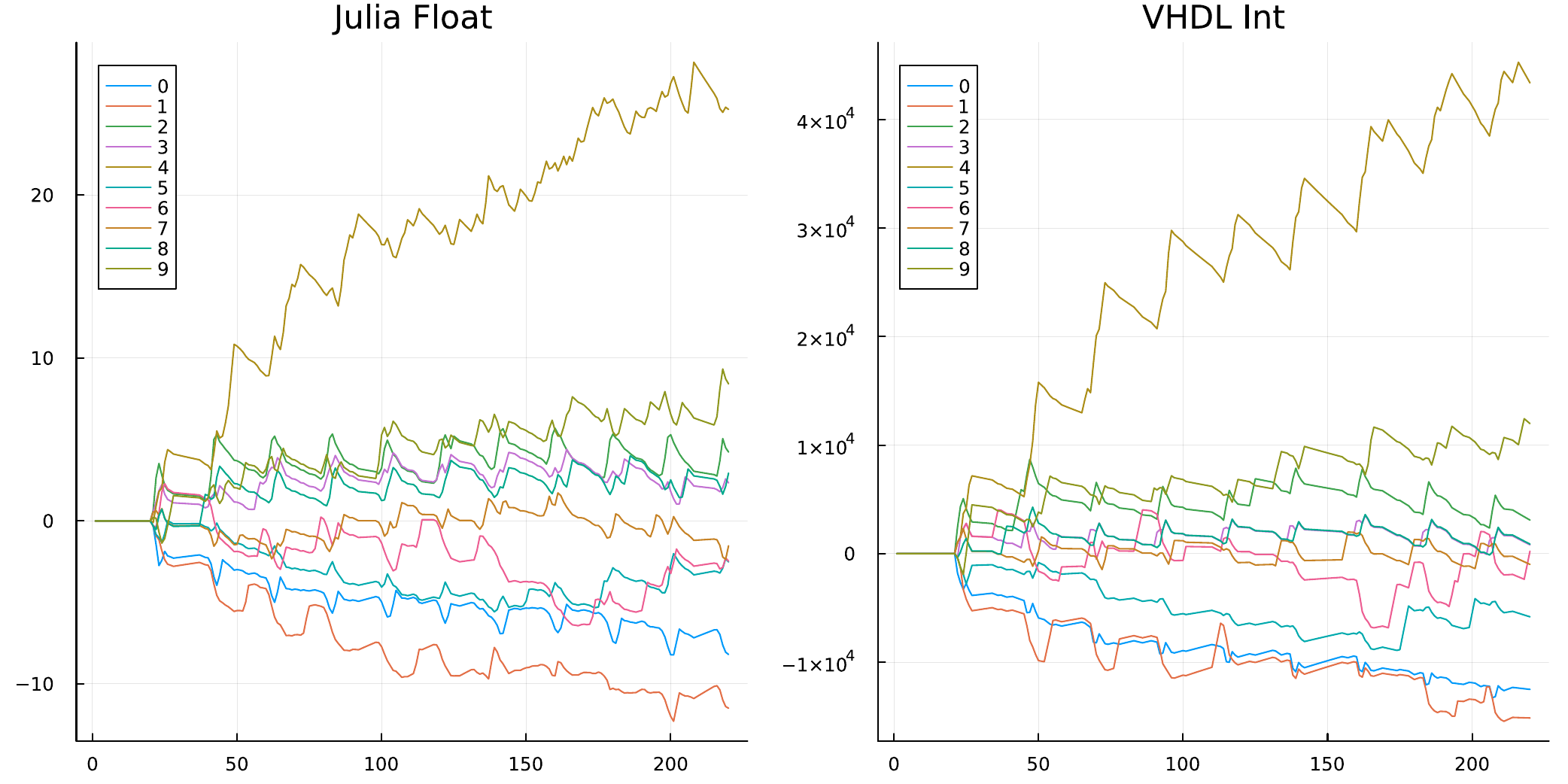} 
    \caption{Stability results and accuracy (the two true solutions).}
    \label{fig:MnistJuliaVsPsVOK}
\end{figure}

\begin{figure}[!t]
    \centering
    \includegraphics[width=15.5cm]{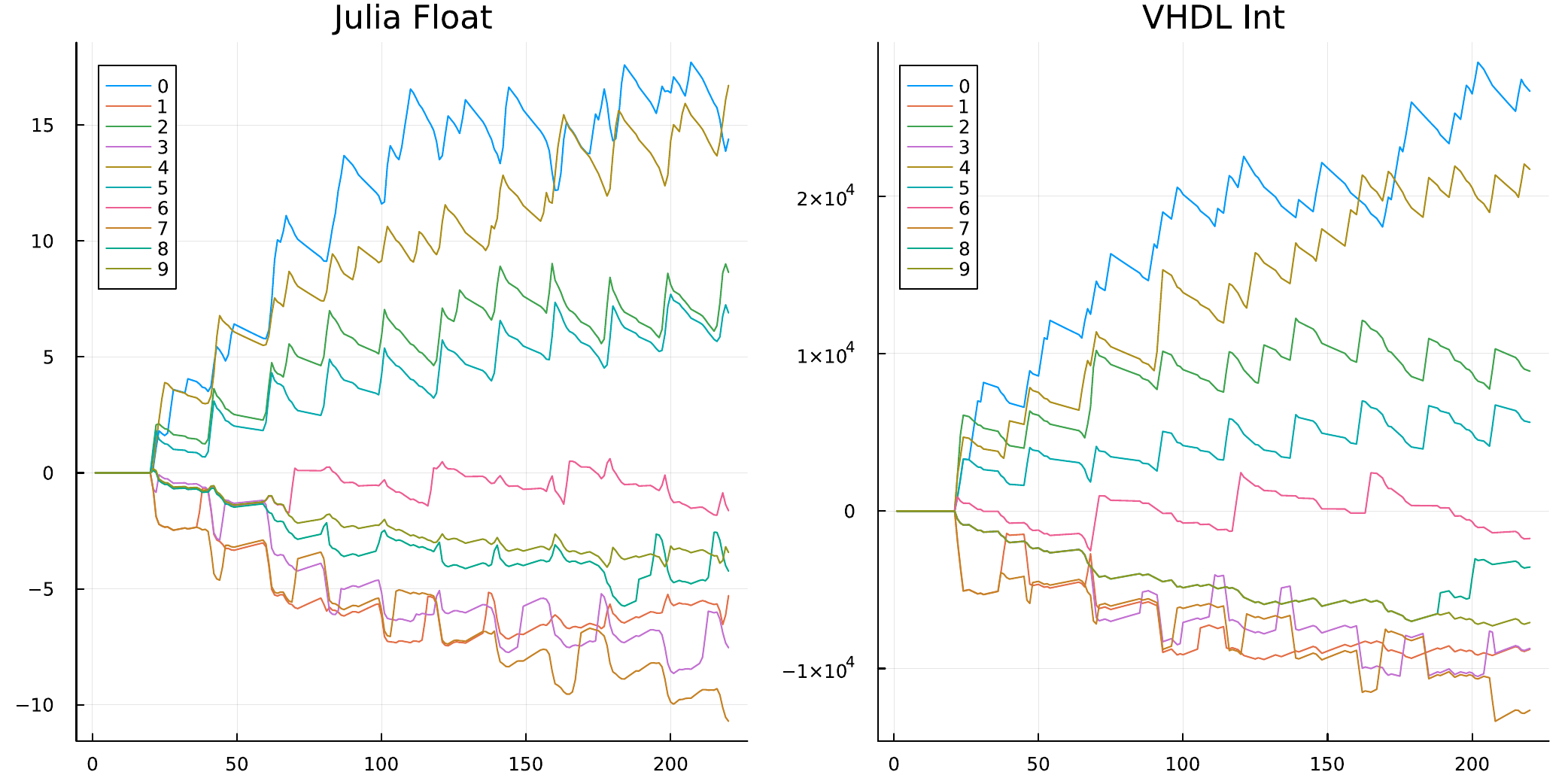}
    \caption{Stability results and accuracy (one true - one false solution).}
    \label{fig:MnistJuliaVsPsVKO}
\end{figure}

Finally, we compare the dynamics of IntegratoR (IR) neuron layer  outputs in both implementations (\cref{fig:MnistJuliaVsPsVOK}). In most cases, both models show very similar output dynamics, the only difference occurs when the classification decision is unclear (\cref{fig:MnistJuliaVsPsVKO}). 

%----------------------------------------------------------------------------- Proposed Architecture
%----------------------------------------------------------------------------- Proposed Architecture
	\begin{figure*}[!t]
		\centering
		\includegraphics[width=15.5cm]{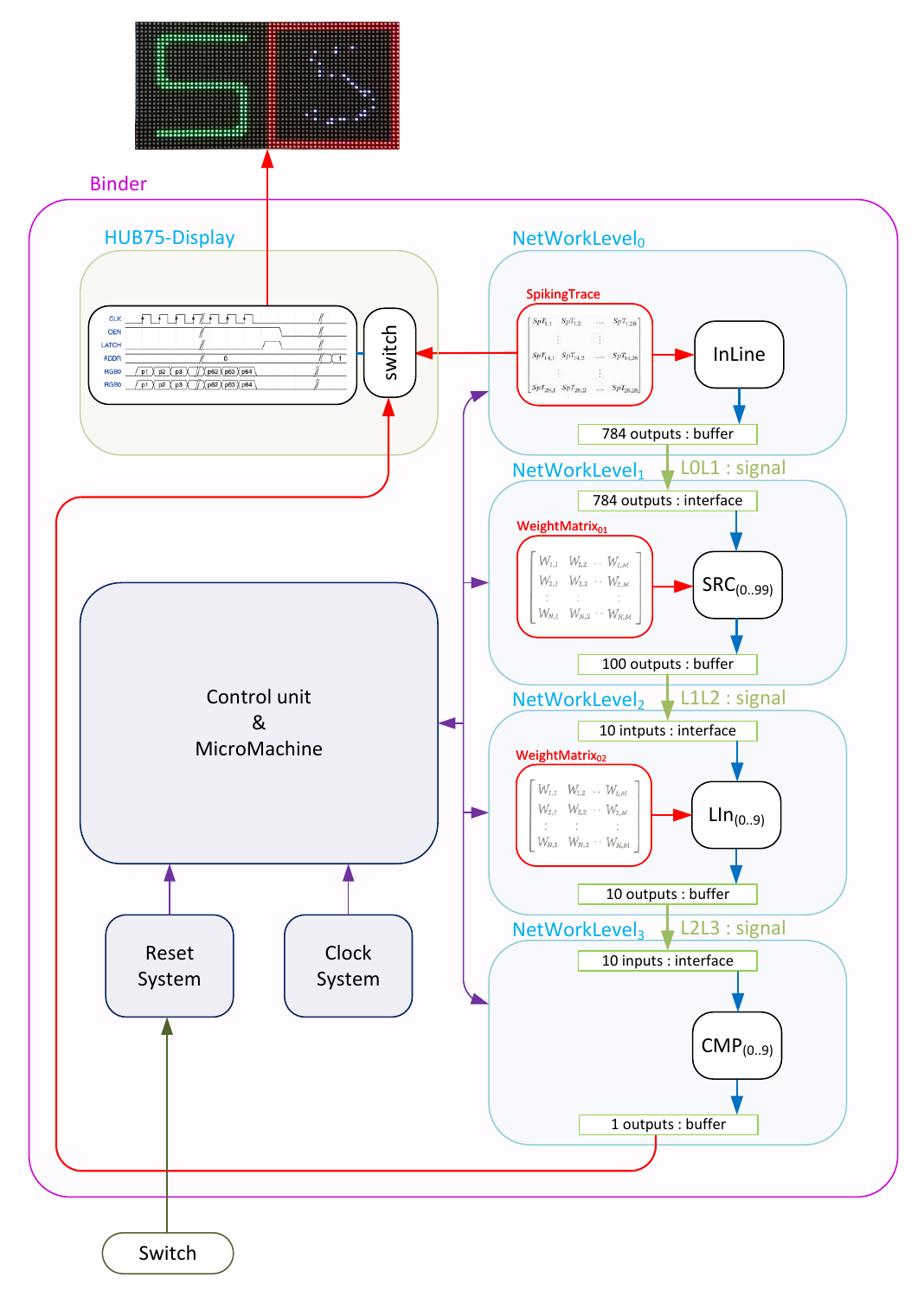}
		\caption{Architecture of the neural network on FPGA.}
		\label{fig:GenGlob01}
	\end{figure*}

\section{Implementation architecture}\label{sec:ProposedArchitecture}
The architecture presented in this work is a layered system, similar to that described in the work of \citet{Carpegna}. Our network is composed of four layers: the input interface (Neuron\_BramInLine\_pkg), the SRC neuron layer (Neuron\_Src2F\_pkg), the IR neuron layer (Neuron\_LIfB\_pkg), and finally a comparator that outputs a value between 0 and 9 (Neuron\_Cmp\_pkg). \Cref{fig:GenGlob01} presents the detailed construction of the architecture. 

In order to obtain an efficient network, all neurons are processed in parallel using the \texttt{GENERATE} operator in VHDL. Since each layer may have a different computation time, synchronization of data transfers between layers is essential. This synchronization is handled by the main module, referred to as the \textit{Binder}. The \textit{Go}, \textit{Ready}, and \textit{Latch} signals ensure synchronization across the different levels. All processes share the same clock (100 MHz) and the same \textit{reset} signal. The \textit{Binder} also routes signals and spike vectors between the different levels.

\begin{figure*}[!ht]
		\centering
		\includegraphics[width=15.5cm]{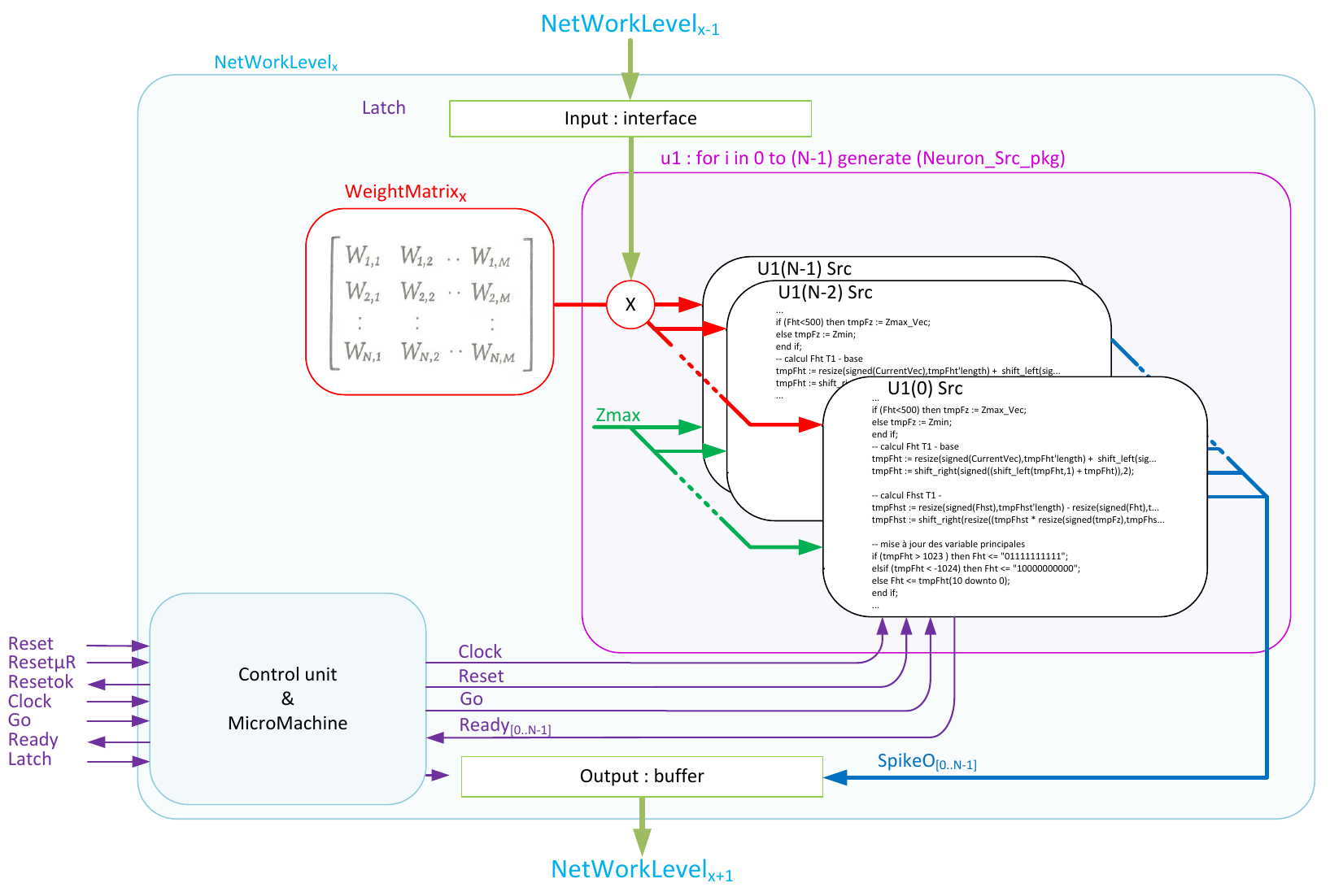}
		\caption{Internal structure of a neural network layer.}
		\label{fig:GenGlob02}
	\end{figure*}

\Cref{fig:GenGlob02} shows the internal structure of each ``NetWorkLevelxx.vhd'' module. They contain a weight matrix encoded in a file named ``WeightMatrixx.vhd'', as well as the code associated with the neuron type used in that layer. To facilitate the layered construction of the network, NetWorkLevel module structures are identical and each module only differ in its internal neuron package.

\subsection{Input interface}
The input layer provides the SRC layer with a vector representing one image from a SpT sequence. Each image is composed of $28 \times 28$ pixels,yielding a vector size of 784 bits. The FPGA board used is an Alinx AX7A200B equipped with an AMD Xilinx Artix-7 XC7A200T shown in \cref{fig:CarteFPGA}. It includes internal BRAM memory in which the SpTs are stored. The NPY files are converted into COE files compatible with Vivado. During this conversion, six additional bits are appended to each image: two bits dedicated to a micro-machine (\textit{\textmu-RESET} and \textit{\textmu-CMP}) and four bits containing the target digit \textit{CMP\_VAL}. These six bits can be considered as neuromodulatory signals. The \textit{\textmu-RESET} signal is activated at the beginning of each SpT to reset the neurons, while \textit{\textmu-CMP} is activated at the end of the trace to enable the comparison between the discriminator output and \textit{CMP\_VAL}.

\subsection{SRC layer and IR layer}
The SRC layer receives a 784-spike input vector, together with two control bits (µ-RESET and µ-CMP), which can be viewed as neuromodulatory spikes, and four additional bits encoding the expected class for final output comparison. It is composed of 100 fully connected SRC neurons. The input current of each SRC is computed following the equation

		\begin{equation}
			I[t] = \beta I[t-1] + \sum_{i=0}^{n} W_{i} Input_{i}
            \label{eqn:CurrentSrvVhdl}
		\end{equation}
		\vspace{0.2mm}

where $n=783$ in our experiments. The weight matrix $W_{ij}$ is imported into the FPGA using a Julia script that generates a file named “WeightMatrix01\_pkg.vhd”, which can be directly used in VHDL. In order to analyze the effect of quantization, the script adjusts the weight bit-width from 9 down to 2 bits (including the sign bit). No additional processing is applied to the weight matrices, in order to validate the robustness of the SRC model. The corresponding results are presented in the following section.

The IR layer receives as input a vector of 100 spikes generated by the SRC layer. The ten IR neurons are fully connected and perform the summation of their inputs as described in \cref{eqn:Lif01}. 

		\begin{equation}
			\smallskip
			S_{out}[t] = S_{out}[t-1] + \sum_{i=1}^{n} \text{Input}(i) \cdot k(i),
			\smallskip
            \label{eqn:Lif01}
		\end{equation}
		\vspace{0.2mm}

Since all weights in the IR weight matrix take values of either $-1$ or $+10$, and in order to limit memory usage, the value $-1$ is encoded as ‘0’ and the value $+10$ as ‘1’. A subsequent \texttt{if} statement is then used to restore the original values. This reduces each IR weight to a single bit. A Julia script exports this matrix into the file “WeightMatrix02\_pkg.vhd”, which is directly usable in VHDL.

The SRC and IR weight matrices are stored in LUT-registers. This choice provides faster access than BRAM, which typically requires at least one clock cycle. This is made possible by the moderate size of the network and the large number of LUTs available on the XC7A200T device.

\subsection{Output layer}
This layer is used for analysis purposes. It takes as input the 10 IR output values and provides the index of the maximum value. At the end of a SpT, this index is compared with \textit{CMP\_VAL} through the activation of the \textit{$\mu$-CMP} signal. In case of a mismatch, an error counter is incremented and subsequently used to compute the recognition accuracy.

\subsection{Display unit - HUB75}
A display unit is implemented to visualize the network behavior during the debugging phase. It uses a $32 \times 64$ LED matrix with a HUB75 interface. The image sequence of a SpT is displayed on the right side, while the recognized digit is displayed on the left side as shown in \cref{fig:GenGlob01}.

\section{Experimental results}\label{sec:Experimentalresults}
\begin{figure*}[p]
		\centering
		\includegraphics[width=16cm]{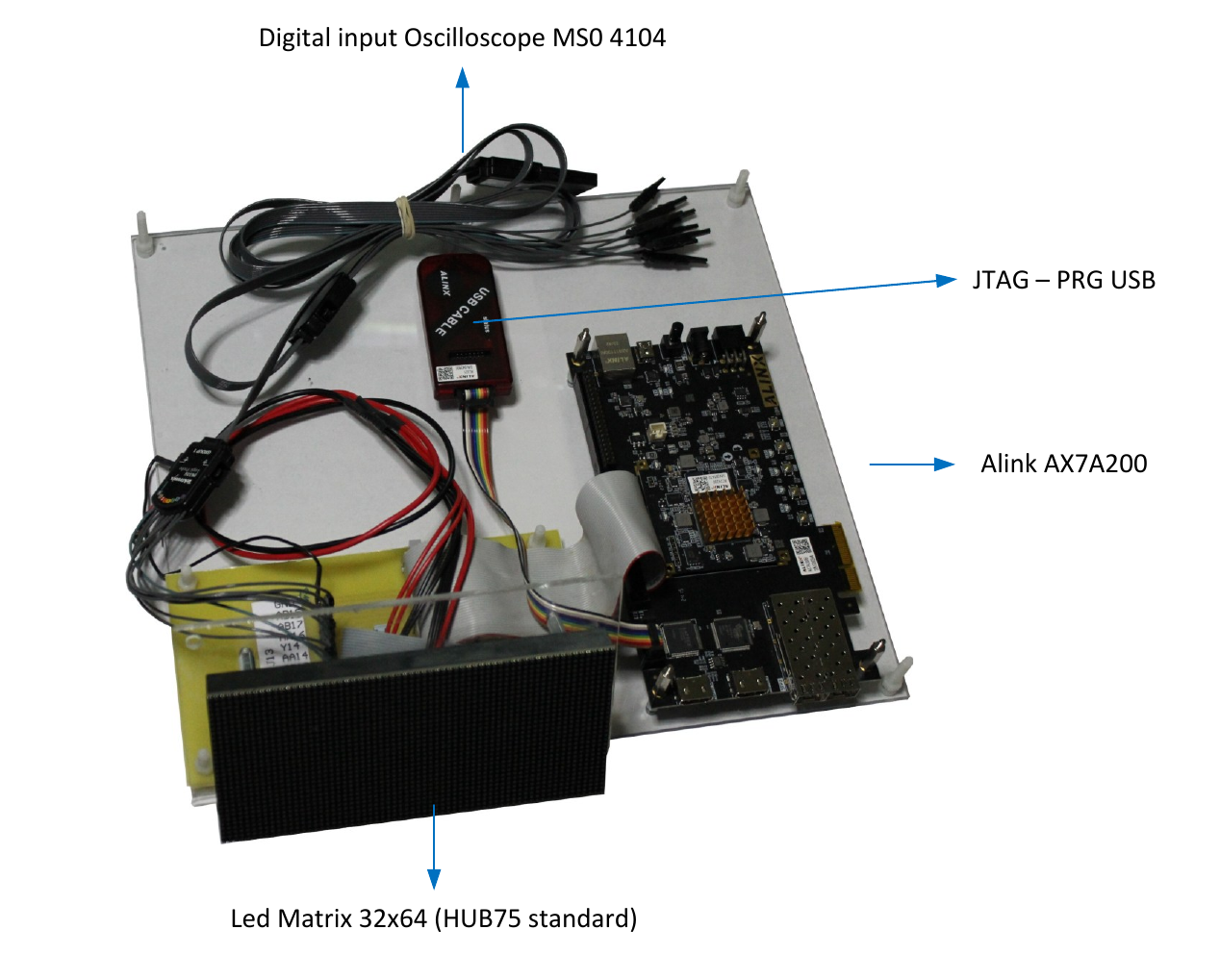}
		\caption{Test setup using the Alinx XC7A200 board.}
		\label{fig:CarteFPGA}
\end{figure*}

\Cref{fig:CarteFPGA} shows the experimental setup used to validate and analyze SRC network performance. All tests were performed on an Alinx board equipped with an AMD Xilinx Artix-7 XC7A200T. All the results presented in this section are obtained using the MNIST dataset, converted into SpTs as described previously.

We first validate the behavior of the VHDL implementation by comparing it with the PsV version coded in Julia. The entire SpT test set is processed on both systems. The parameters $z^{hyp}_{s}$ and $z^{deep}_{s}$ are set to 900 and 100, respectively, in both cases. The SRC input weight matrices are encoded using 9 bits (from $-256$ to $+255$). The VHDL implementation deployed on the FPGA achieves an accuracy of 96.31\%, which is identical to that of the PsV implementation in Julia (96.31\%).

\begin{figure}[h]
    \centering
    \begin{subfigure}[c]{0.45\linewidth} % ou mettre b pour bottom
        \centering
        \includegraphics[angle=90,width=\linewidth]{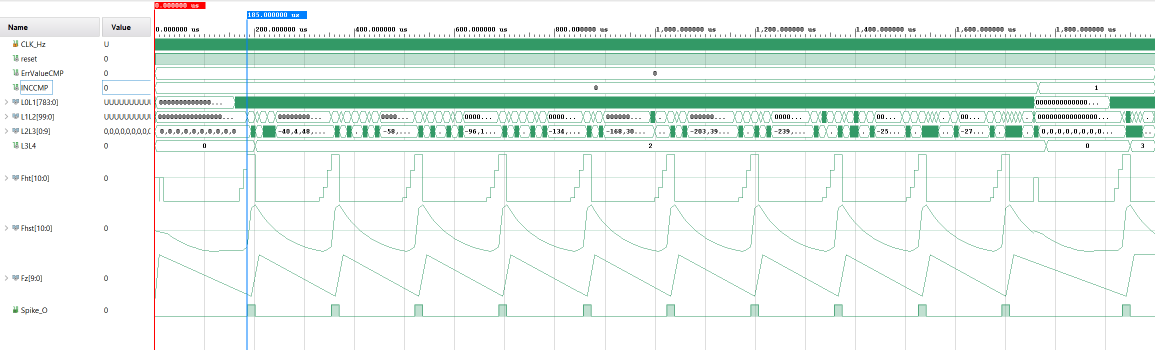}
        \caption{}
        \label{fig:Tlatence_a}
    \end{subfigure}\hfill
    \begin{subfigure}[c]{0.55\linewidth} % ou mettre b pour bottom
        \centering
        \includegraphics[angle=90,width=\linewidth]{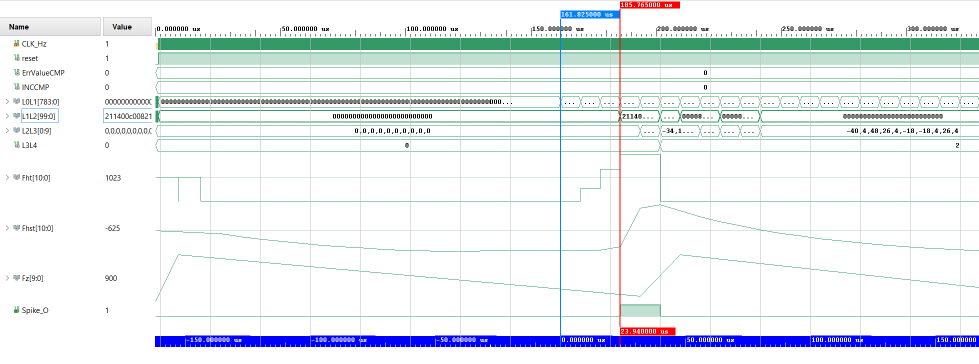}
        \caption{}
        \label{fig:Tlatence_b}
    \end{subfigure}

    \caption{(a) For one input sample from the MNIST database, the first spike in an SRC neuron appears after 185 µs.\\
    \centering{(b) Input-to-activation latency of an SRC neuron.}}
    \label{fig:Tlatence}
\end{figure}

We then analyze the computation time and the hardware resource utilization of the FPGA. The computation time directly influences the energy consumed to process each digit. A digit corresponds to a SpT of 220 images. The SRC layer is the slowest layer of the network: each SRC neuron requires 784 clock cycles to process a single image. At the end of each image processing step, 8 additional clock cycles are required for buffering and for resetting the state machines. This results in a total of 174,240 clock cycles per digit. At a clock frequency of 100 MHz, this corresponds to a processing time of 1.7424 ms per digit. \Cref{fig:Tlatence_a} illustrates the
complete processing flow of a single image, including the latency between the start of the SpT sequence and the first activation of an SRC cell (185 $\mu$s in this experimental example). \Cref{fig:Tlatence_b} shows the latency between the arrival of an input spike at an SRC neuron and the subsequent activation of this neuron, measured here as 25 $\mu$s.

The processing time is then reduced by shortening the SpT length, and the resulting degradation in accuracy is evaluated. \Cref{tab:res1} shows a significant degradation for sequences shorter than 44 images.

 	\begin{table}[h]
			\caption{SpT length vs. accuracy and energy efficiency at 100 MHz.}
			\label{tab:res1}
			\centering
			\begin{tabular}{@{}l c c c}
			\toprule
			size  & Accuracy (\%) & $T$/SpT (ms) & E/SpT (mJ)\\
			\midrule
			20+200   &  96.31 & 1.748 & 1.972\\
			10+100   &  95.66 & 0.871 & 0.985\\
			5+50     &  95.25 & 0.436 & 0.492\\
			4+40     &  94.17 & 0.348 & 0.394\\
			2+20     &  91.99 & 0.174 & 0.197\\
			\bottomrule
			\end{tabular}
 	\end{table}

Finally, the energy per digit can be estimated from Vivado power estimation (post-implementation). Vivado reports the power consumption of the implemented logic after compilation. The total reported consumption is 1.13 W (including the HUB75 display). The last column of \cref{tab:res1} provides the energy required to process a single digit, expressed in mJ. To avoid an excessive degradation of accuracy, the following tests are not conducted using sequences shorter than 44 images.

Hardware resource utilization can also contribute to lowering energy consumption. Vivado provides a breakdown in terms of LUTs, FFs, BRAMs, and DSPs. \Cref{tab:SizeLUT} reports the FPGA hardware resource utilization required for the implementation of different neuron models on an Artix-7 device, as presented in \citet{koravuna2024spiking}, and compares them with the proposed SRC model. The results clearly demonstrate the low hardware cost associated with the implementation of an SRC neuron.

	\begin{table}[t]
			\caption{FPGA Resource Utilization by Neuron Model Type.}
			\label{tab:SizeLUT}
			\centering
			\begin{tabular}{@{}l r r r@{}}
			\toprule
			Neuron type & LUT & FF & DSP\\
			\midrule
			Leaky Integrate-and-Fire (LIF) \cite{koravuna2024spiking}  & 13  & 17 & 0  \\
            SRC (ours) & 75  & 21 & 0 \\
            Quadratic Integrate-and-Fire (QIF) \cite{koravuna2024spiking}  & 82  & 21 & 0  \\
			Izhikevich Model  \cite{koravuna2024spiking} & 42  & 25 &  1\\
			Hodgkin-Huxley (HH) \cite{koravuna2024spiking}   & 73  &  25 & 3\\
    		\bottomrule
			\end{tabular}
 	\end{table}

\subsection{Binary quantization}
\Cref{tab:sizeReg} presents the results obtained by varying the register size of the SRC weight matrix. The maximum size is 9 bits, and the minimum size is 2 bits (including the sign bit). A significant degradation is observed when the weight register size is less than or equal to 3 bits. In contrast, the use of 4-bit weights represents an attractive trade-off: accuracy remains acceptable while LUT utilization is reduced by approximately 5 to 6\% , which should lead to a proportional reduction in power consumption due to the released logic resources. The results also show that the spike generation rate can be reduced by increasing $z^{hyp}_{s}$ (\cref{fig:HsvsZdeep}).

 	\begin{table}[h]
			\caption{Neural network accuracy as a function of the matrix weight bit-width and the  $z^{hyp}_{s}$ parameter value.}
			\label{tab:sizeReg}
			\centering
			\begin{tabular}{@{}l c c c c c c c}
			\toprule
			Matrix            & acc. (\%)        & acc. (\%)         &  acc. (\%)          & acc. (\%)          & acc. (\%)          &  acc. (\%)       &  acc. (\%)      \\
            weight            & $z^{hyp}_{s}$=880& $z^{hyp}_{s}$=900 & $z^{hyp}_{s}$=920   & $z^{hyp}_{s}$=940  & $z^{hyp}_{s}$=960  & $z^{hyp}_{s}$=980&  $z^{hyp}_{s}$=1000 \\
            bit-width         &                  &                   &                     &                    &                    &                  &                 \\
			\midrule
			9                 & 96.13            & 96.31             &  96.26              & 96.28              & 96.02              & 95.58            & 93.23          \\
			8                 & 96.17            & 96.25             &  96.21              & 96.31              & 96.07              & 95.45            & 92.98          \\
			7                 & 96.06            & 96.27             &  96.21              & 96.30              & 96.07              & 95.52            & 92.72          \\
			6                 & 96.15            & 96.26             &  96.26              & 96.24              & 95.88              & 95.22            & 92.18          \\
			5                 & 96.04            & 96.14             &  96.35              & 96.13              & 95.73              & 94.43            & 90.64          \\
			4                 & 95.17            & 95.35             &  95.40              & 95.36              & 93.75              & 91.60            & 86.17          \\
   			3                 & 90.72            & 91.01             &  90.69              & 89.94              & 86.00              & 81.37            & 73.16          \\
            2                 & 55.01            & 54.82             &  53.74              & 51.48              & 47.61              & 43.67            & 37.58          \\
			\bottomrule
			\end{tabular}
 	\end{table}

The reference value of $z^{hyp}_{s}$ is set to 900 in order to closely match the behavior reported by \citet{DeGeeter}. However, higher values (e.g., 980) significantly reduce spiking activity (by approximately a factor of $\sim$3 between 880 and 980 in our observations). This suggests that an \textit{event-based} implementation could become advantageous when activity is sufficiently sparse.

We then analyze recognition accuracy when weight quantization is combined with a reduction in SpT length, for $z^{hyp}_{s}$ values ranging from 880 to 980. The results are reported in \cref{tab:SpTVsReg}. When SpT length becomes smaller and weights are quantized on 4 bits, we start to see a decrease in accuracy for low spiking frequencies (i.e. higher $z^{hyp}_{s}$ values). 

 	\begin{table}[ht]
			\caption{SpT vs Size Reg. vs $z^{hyp}_{s}$}
			\label{tab:SpTVsReg}
			\centering
			\begin{tabular}{@{}r | r r r | r r r | r r r }
			\toprule
	                & \multicolumn{3}{c}{880} & \multicolumn{3}{c}{900} & \multicolumn{3}{c}{920} \\
                     \cline{2-10}
             Size SpT  & 6 bits & 5 bits & 4 bits  & 6 bits  & 5 bits & 4 bits & 6 bits & 5 bits & 4 bits \\
			\midrule
			       55   &95.13\% & 95.05\%&  94.38\%&  93.81\%& 93.32\%& 91.90\%& 94.26\%& 93.68\%& 92.48\%\\
			       44   &94.07\% & 93.66\%&  92.39\%&  94.39\%& 93.91\%& 92.89\%& 94.87\%& 94.34\%& 93.44\%\\
			\addlinespace \midrule \addlinespace
	                & \multicolumn{3}{c}{940} & \multicolumn{3}{c}{960} & \multicolumn{3}{c}{980} \\
                     \cline{2-10}
            Size SpT  & 6 bits  &5 bits  & 4 bits  & 6 bits  & 5 bits & 4 bits & 6 bits  &5 bits & 4 bits \\
			\midrule
			       55   &94.78\% & 94.17\%&  92.99\%&  90.24\%& 89.02\%& 85.35\%& 90.74\%& 89.97\%& 86.46\%\\
			       44   &91.10\% & 89.78\%&  87.05\%&  90.71\%& 89.49\%& 86.63\%& 91.40\%& 90.72\%& 88.17\%\\
			\bottomrule
			\end{tabular}
 	\end{table}

\subsection{Comparison with other SNN FPGA implementations}
\begin{table}[ht]
			\caption{Comparison of this SRC implementation with FPGA-based SNNs.}
			\label{tab:Workcmp3}
			\centering
			\begin{tabular}{@{}l c c c c }
			\toprule
 			Design          & Carpegna et al. \cite{Carpegna2} & Han et al. \cite{Han}&  this work (200 - 20)  &  this work (44 - 4) \\
 			\midrule
			Year            &  2024                            &     2020            &   2025                &   2025       \\
			Model           &  LIF                             &    LIF              &   SRC                 &    SRC       \\
            Neurons         &  784-128-10                      &   784-1024-1024-10  &   784-100-10          &  784-100-10  \\
            FPGA            &  XC7Z020                         &  C7Z045 SoC         &   XC7A200             &  XC7A200     \\
            $freq.$ [MHz]   &  100                             &  100$^1$            &   100                 &  100         \\
            Used Logic cells&  7,612                           &  12,690             &   -                   &  93,347      \\
            Used DSP        &  0                               &  -                  &   100                 &  100         \\
            Used BRAM       &  18                              &  40.5               &   341$^2$             &  341$^2$     \\
			Weight (bits)   &  4                               &  6    (4)           &   4                   &  4           \\
            $T$/img [ms]    &  0.78                            &  12.42$^1$          &   1.748               &  0.436       \\
            E/img [mJ]      &  0.14                            &  5.92$^1$           &   1.972                &  0.394        \\
            Training accur. [\%] &   96.83                     & 98.48               &   96.48               &  96.48       \\
            Accuracy [\%]   &  93.85                         &  97.06  ($<$20)     &   95.35               &  92.89       \\
            
			\bottomrule
            \multicolumn{5}{l}{$^1$ adjust from 150 to 100 MHz  } \\
            \multicolumn{5}{l}{$^2$ BRAM is used for storing the SpTs.} \\
	       	\end{tabular}
            \vspace{2mm}
 \end{table}

Our results are compared with other FPGA-based SNN implementations evaluated on the MNIST dataset. To ensure a fair comparison, the operating frequency is normalized to 100 MHz. Since the neuron model proposed in this work (SRC) differs from conventional approaches, the comparison is conducted with studies employing neurons of comparable complexity (LIF and variants) as well as neural networks of similar size.

For comparable weight matrix word widths (4 bits), comparing the accuracy gaps between the training and test datasets across the different works reveals a discrepancy of 3\% in the study by \citet{Carpegna2} and more than 70\% by \citet{Han}, which is significantly higher than the 1.1\% observed in this work. This strong preservation of accuracy after FPGA deployment could be attributed to the robustness of the underlying neuronal model, which implements spiking dynamics without reset through a mixed feedback interconnection.

The processing time per digit is below one millisecond for short SpTs and is comparable to that reported by \citet{Carpegna2}, while the energy consumption is particularly low. This performance is primarily due to extensive operator simplifications and full parallelization within each layer. As a result, SRC neurons are able to compete with LIF neurons in terms of computation time and energy consumption, while preserving continuous dynamics that are closer to those of biological neuron models, as well as analog electronic implementations \cite{indiveri2025neuromorphic,mendolia2025neuromodulable}. 

\subsection{Modularity Evaluation of the Architecture}
Finally, we demonstrate the modularity and robustness of the proposed architecture by testing the performance of a multi-layer SRC network on the Fashion-MNIST dataset \citet{xiao2017FashionMNIST}. This network consists of a 784-input layer, four layers of 100 SRC neurons, a 10-neuron IR layer, and finally an output comparator layer. As in the previous network, training was performed offline using the original SRC equations and weight matrices were directly transferred on the FPGA. \Cref{tab:Fashion-MNIST} reports the results obtained in software and hardware with our network and compare with those reported by \cite{mirsadeghi2026multiplication}.

\begin{table}[h]
			\caption{Accuracy comparison of different SNN methods on Fashion-MNIST.}
			\label{tab:Fashion-MNIST}
			\centering
			\begin{tabular}{@{}l c c }
			\toprule
 			Method                                         & architecture                      & accuracy (\%) \\
 			\midrule
            IF (Q5.7)  \cite{mirsadeghi2026multiplication}      &  784-600-600-10                   &   84.8       \\
            SRC (Float - julia) & 784-100-100-100-100-10 &   88.8       \\
            SRC (bit-width 9) &  784-100-100-100-100-10 &   83.2       \\
            SRC (bit-width 6) &  784-100-100-100-100-10 &   80.3       \\
            SRC (bit-width 5) &  784-100-100-100-100-10 &   72.2       \\
            SRC (bit-width 4) &  784-100-100-100-100-10 &   43.4      \\   
			\bottomrule
	       	\end{tabular}
            \vspace{2mm}
\end{table}

Results show that network architecture of the FPGA implementation can be easily modulated, and that the FPGA implementation can maintain high accuracy on more complex benchmarks. However, the performance degradation between software and hardware is noticeably higher than for the single layer network. This important difference can be explained by the accumulation of numerical errors across the successive layers of the network. The SRC implementation however achieves similar results to those reported in the literature \citet{mirsadeghi2026multiplication} on the same FPGA family, Artix-7 device.

The modularity of the proposed network could enable the use of multiple interconnected FPGA devices, although this step has not been experimentally validated. In such configuration, the different network layers would be distributed across the FPGA devices. It would however be necessary to transmit the spike vector from a given NetWorkLevelx module to a NetWorkLevely module, i.e., from one FPGA to another. This would be feasible provided that the size of the spike vector does not exceed the number of physical input/output pins available on the FPGA chip. This constraint could be relaxed if, as presented in  \citet{liang2025quantitative}, spike transmission were performed through a transceiver. The use of such a mechanism would require the addition of a serial controller, clocked and synchronized by the Binder layer. \citet{liang2025quantitative} demonstrate that large-scale neural networks comprising more than 500,000 neurons can be executed on multi-FPGA clusters, such as a 48-FPGA platform, by exploiting FPGA transceivers for inter-device communication.

\section{Discussion}\label{sec:discussion}
The experiments highlight two main trade-offs. First, reducing the length of the SpTs linearly decreases the processing time and energy consumption, but the accuracy degrades when the sequence length falls below approximately 44 images (\cref{tab:res1}). Second, weight quantization reduces the usage of logic resources (and thus potentially the power consumption), but the accuracy degrades sharply when the word width is reduced below 4 bits for MNIST database (\cref{tab:sizeReg}). In the MNIST configuration, 4-bit synaptic weights therefore appear to be a practical lower bound, since they preserve a high level of accuracy while reducing the hardware cost. However, the Fashion-MNIST experiment shows a much stronger sensitivity to quantization, with a significant accuracy drop when the bit-width is reduced to 4 bits. This suggests that the optimal quantization level depends not only on the neuron model, but also the  neural network architecture and the task itself. As the number of layers increases, approximation and quantization errors may accumulate across successive processing stages, potentially leading to a progressive degradation of the final classification accuracy. This is also a feature of our experimental approach. We relied on hardware-agnostic training using the original SRC equations, followed by a direct transfer of the weight matrices into the FPGA implementation. This strategy serves as a strict validation of the architecture inherent robustness, proving that the simplified hardware-mapped dynamics remain stable without hardware-in-the-loop optimization. Nonetheless, incorporating hardware-aware training paradigms or on-chip learning mechanisms represents a promising avenue for future work that could further optimize classification performance and mitigate quantization constraints.

The SRC model also introduces an important control parameter $z^{hyp}_{s}$, which directly regulates the spike generation frequency (\cref{fig:HsvsZdeep}). We observe that increasing $z^{hyp}_{s}$ can reduce spiking activity by a factor of approximately three (from 880 to 980) while maintaining an acceptable level of accuracy. This suggests that an \textit{event-driven} scheduling strategy could become advantageous when the spiking activity is sufficiently sparse. A natural extension of this work is therefore to investigate a hybrid approach capable of switching between synchronous processing and \textit{event-driven} execution depending on the observed spike density.

Finally, beyond reporting energy consumption per digit or per image, it may be more informative to normalize energy with respect to the number of emitted spikes. A metric such as $E/\text{spike}$ could facilitate comparisons between implementations of different sizes and operating frequencies. This would require measuring the total number of spikes produced (e.g., at the SRC outputs and/or across the network layers) and reporting $E/\text{spike}$ in addition to $E/\text{digit}$.

\section{Conclusion}\label{sec:conclusion}
In this work, we implemented a new SRC neuron model on SNNs, and evaluated it on an Artix-7 XC7A200T FPGA using SpTs derived from the MNIST dataset and Fashion-MNIST. The SRC neuron preserves continuous dynamics while remaining compatible with efficient hardware simplifications in VHDL. We develloped a \textit{Binder--NetWorkLevelxx} architecture that simplifies reconfiguration, since the network weight matrices can be directly loaded from the off-line training results. This allows for neural networks comprising a large number of layers to be implemented according to the \textit{Binder--NetworkLevelxx} design principle.

Our prototype achieves $96.31$\% accuracy with a 220-image SpT and a processing time of $1.7424$ ms per digit at $100$ MHz. We further show that reducing the SpT length and quantizing the synaptic weights can significantly lower energy consumption, achieving 93.32\% accuracy at 0.492 mJ per digit (55 images, 5-bit weights) and 92.89\% at 0.394 mJ (44 images, 4-bit weights). These results indicate that SRC neurons are strong candidates for energy-efficient SNN accelerators while requiring low hardware resource utilization. However, deeper networks may accumulate approximation and quantization errors across successive processing stages, potentially reducing the final classification accuracy. This point should be considered when extending the proposed approach to larger architectures.

Beyond the reported performance results, the main contribution of this work is to provide a practical hardware design methodology for SRC-based SNNs. The article gives the reader a complete implementation path, including mathematical simplification of the neuron equations, VHDL implementation, modular network construction, direct loading of off-line trained weights, and quantitative evaluation of the trade-offs between accuracy, latency, energy consumption, and FPGA resource utilization.

Future work includes completing the combined exploration of SpT length and weight quantization, investigating adaptive \textit{event-driven} scheduling enabled by reduced spiking activity, and developing automation tools for code generation and interconnection to support larger and potentially multi-FPGA networks.

\clearpage

\clearpage

\clearpage

%\section*{References}
\bibliography{mybib.bib}{}
\bibliographystyle{unsrtnat}

\end{document}